\documentclass[conference]{IEEEtran}
\IEEEoverridecommandlockouts 
\usepackage{fancyhdr} 
\usepackage{url}
\usepackage{cite}
\usepackage{amsmath,amssymb,amsfonts}
\usepackage{algorithm}
\usepackage{algpseudocode}
\usepackage{graphicx}
\usepackage{textcomp}
\usepackage{xcolor}
\def\BibTeX{{\rm B\kern-.05em{\sc i\kern-.025em b}\kern-.08em
    T\kern-.1667em\loIr.7ex\hbox{E}\kern-.125emX}}

\title{AI-driven Transformer Model for Fault Prediction in Non-Linear Dynamic Automotive System}
\author{
    Priyanka Kumar\thanks{This work was supported by ONR/Grant Number - N000014-21-1-233.} \\
    Department of Computer Science \\
    The University of Texas at San Antonio\\
    San Antonio, Texas, United States\\
    \texttt{priyanka.kumar@utsa.edu}
}
\begin{document}
\maketitle

\begin{abstract}
Fault detection in automotive engine systems is one of the most promising research areas. Several works have been done in the field of model-based fault diagnosis. Many researchers have discovered more advanced statistical methods and algorithms for better fault detection on any automotive dynamic engine system. The gas turbines/diesel engines produce highly complex and huge data which are highly non-linear. So, researchers should come up with an automated system that is more resilient and robust enough to handle this huge, complex data in highly non-linear dynamic automotive systems. Here, I present an AI-based fault classification and prediction model in the diesel engine that can be applied to any highly non-linear dynamic automotive system. The main contribution of this paper is the AI-based Transformer fault classification and prediction model in the diesel engine concerning the worldwide harmonic light vehicle test procedure (WLTP) driving cycle. This model used 27 input dimensions, 64 hidden dimensions with 2 layers, and 9 heads to create a classifier with 12 output heads (one for fault-free data and 11 different fault types). This model was trained on the UTSA Arc High-Performance Compute (HPC) cluster with 5 NVIDIA V100 GPUs, 40-core CPUs, and 384GB RAM and achieved 70.01\% accuracy on a held test set.
\end{abstract}


\section{Introduction}
 Diesel engines have been used in naval vessels for many years due to their reliability, durability, and efficiency. It is also used in auxiliary systems on board naval vessels, such as electrical generators and pumps. So these engines play a critical role in maintaining the operational readiness of naval vessels and are an essential component of modern naval warfare. In the modern era, misdiagnosis and robustness are major concerns in any automotive vehicular system. Research challenges in the current systems include:
\begin{enumerate}
    \item Robustness is one of the major concerns in any automotive or diesel engine system for critical applications. The reason behind this is that many manufacturers make their products and sell their vehicles all over the world. So, it is a challenging problem for manufacturers to come up with one resilient fault detection method with respect to wide variations in weather, different driving styles, heavy traffic conditions, etc.
    \item With the advancement of technologies, the development of autonomous vehicles/engines is a special concern. The onboard computer needs to keep track of the engine's health without the aid of a human operative. So, it is of utmost importance to come up with a self-aware system that will take care of their occupant’s life and health by self-diagnosis and self-healing continuously.
    \item These automotive engines produce huge data that is complex in nature and difficult to handle. So, the Internet of Things, allowing for edge computing and requiring less energy to operate is critical for exploratory vehicles.\\

With these motivations, I designed a Transformer model that is beneficial in fault classification and prediction for diesel engines due to their advanced capabilities in handling sequential data and capturing long-range dependencies. Due to self-attention mechanisms, it focuses on the different parts of the sequences and on critical aspects of the data that are most indicative of faults. This is especially useful for identifying subtle and complex patterns in engine sensor data that could signal impending faults. Diesel engines produce a multitude of sensor data streams. Transformers are adept at processing multivariate time series data, making them suitable for analyzing the various types of signals generated by engine sensors. 
\end{enumerate}

\section{Background}
In most automotive systems, diagnostic systems monitor various components of the engine system and are independent of each other. The purpose of the system is then to map the changes in the sensor data to various fault types that may or may not be dependent on each other. Due to the dependent nature of the faults, a highly intelligent algorithmic framework is needed to isolate individual faults. Previous
methodologies have the major disadvantage of not being able to detect faults in chronological order \cite{SILVA201210977, scacchioli2006model}. Some systems reconfigure the vehicle to operate at reduced performance levels to avoid fault prediction using this conservative approach \cite{6710264, 6144754, 4711448}. In 2022, authors introduced an adaptive sparse attention network that enhances fault detection by focusing on decentralized local fault information in real-time \cite{Jiang22}. The introduction of a new soft threshold filter that improves the visualization and interpretation of fault mechanisms. This approach significantly enhanced the interpretability of the diagnostic process compared to traditional deep learning models. However, a more robust AI model is required which will classify huge and complex patterns of faulty and non-faulty classes with their types and behaviors.

Xisheng Jia et al. designed a novel method for diagnosing faults in diesel engines by combining optimized Variational Mode Decomposition (VMD) with Deep Transfer Learning (DTL) \cite{electronics11131969}. By fine-tuning a pre-trained ResNet18 model on ImageNet samples, the approach successfully reduced noise and improved diagnostic efficiency and accuracy, minimizing the need for manual feature extraction and expert experience. Raj et al.’s work presents a novel framework that utilizes advanced machine learning algorithms and big data analytics to enhance the predictive maintenance of turbines \cite{raj2019towards}. By accurately predicting potential failures and optimizing maintenance schedules, the framework significantly improves turbine reliability and operational efficiency while reducing downtime and costs ~\cite{balakrishnan2021stochastic}.

In this research work, the proposed Transformer model classifies the faults and predicts the faults in chronological order.
\begin{figure*}[ht]
    \centering
    \includegraphics[width=0.9\textwidth]{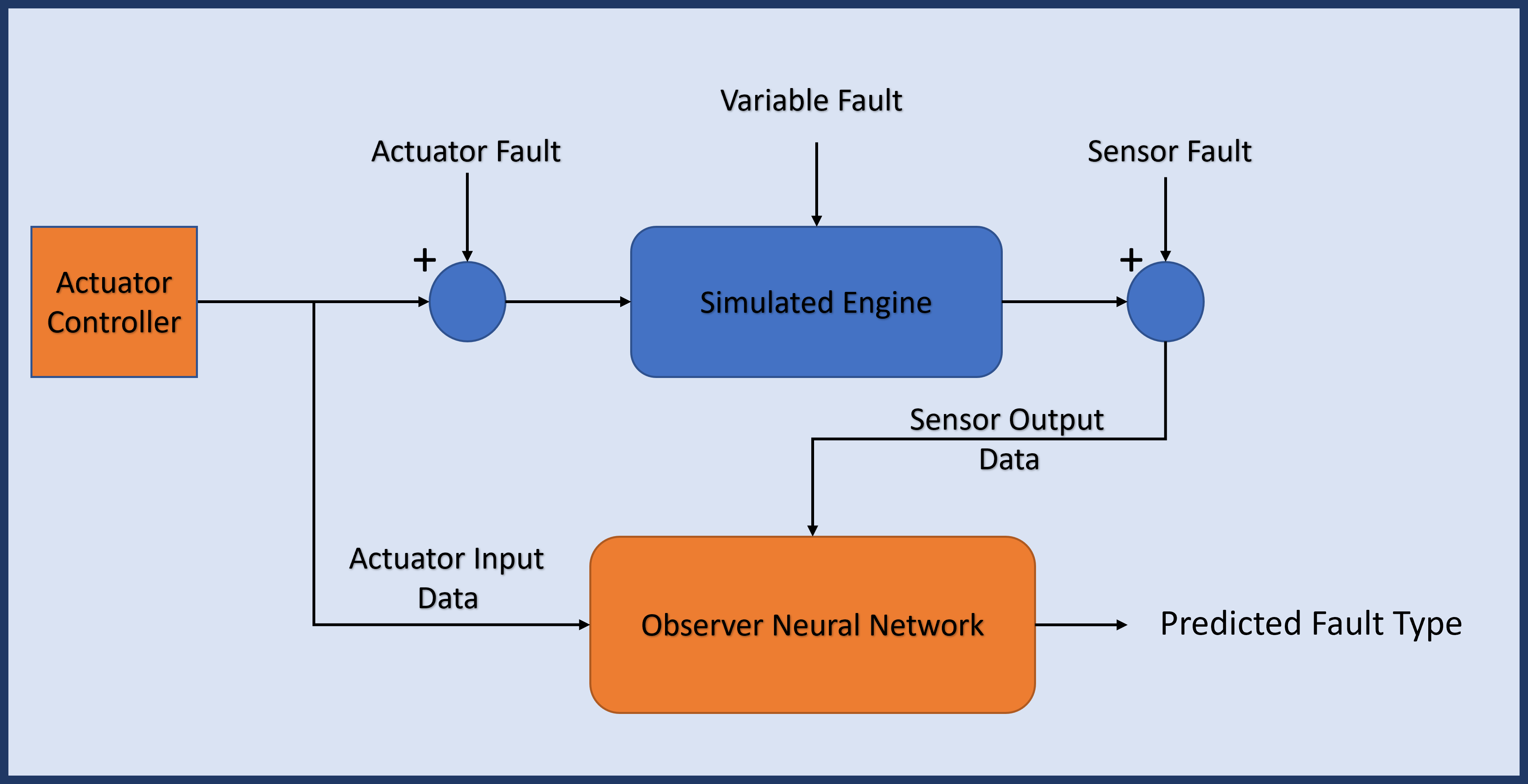}
    \caption{Diesel engine error detection}
    \label{fig:our_approach}
\end{figure*}
\section{The proposed framework}
With the background study and literature survey, considering the current research challenges in the field of automotive engine systems I propose a new framework that will create and perform model-based fault diagnosis which will be a more effective design to demonstrate and analyze fault detection and fault prediction. This can be tested for any prototype application and then finally deployed it into the actual engine system. The proposed framework consists of two parts:
\begin{enumerate}
    \item Replace the Engine Observer: Previous methods~\cite{Ng2020} use a physics engine to simulate the internal states of the real-world engine and compare the simulated sensor outputs with the observed outputs to create a function to map the differences, called residuals, to fault types. I intend to replace this computationally complex task with a Neural Network that can perform the mapping of the inputs and outputs of the real engine to the fault types.
    \item Early Fault Detection: Instead of reading current faulty sensor data to identify the fault type, I also want to predict the fault occurring before the damage to the engine system occurs. For this our Neural Network approach keeps an internal history of the previous sensor outputs to predict a faulty sensor output before it happens.
\end{enumerate}

\subsection{Objectives}

1. Create a Neural Network (NN) based generator for these normal data and then faulty data that are more robust and resilient than the existing one. The main focus will be the Transformer model.

2. Then classify the fault-free and fault-free data, including the type of fault.

3. The model should be able to detect a fault as soon as it occurs using time series data.

\subsection{Test Environment - UTSA ARC High-Performance Computing Cluster}
This research work was tested on the UTSA Arc HPC cluster, a high-performance computing environment.  I have used 2 GPU nodes each containing two CPUs, 4 V100 GPUs, and 384 GB RAM. I have also tested the Transformer model on Arc's large memory nodes, one of which was equipped with two AMD EPYC CPUs and 2 TB of RAM and another node equipped with two AMD EPYC CPUs and 1 TB of RAM.  I observed the total time taken for the training and the time required for each epoch.  After experimenting on the GPU and AMD CPU, I observed there is no significant difference in accuracy and time taken between each epoch during the training of the model.

\subsection{Approach}
I have used a 397 GB dataset generated by the TCSISimTestbed simulator using Matlab/Simulink \cite{Ng2020}. The dataset is currently hosted on the UTSA Arc HPC environment for experimental purposes. This dataset was generated using the Worldwide Harmonized Light Vehicle Test Procedure cycle among the 4 existing types of standard driving cycle, the Worldwide Harmonized Light Vehicle Test Procedure, the New European Driving Cycle, the Extra-Urban Driving Cycle, and the US Environmental Protection Agency Federal Test Procedure \cite{9036118}. Using WLTP, I have used a fault-free dataset and a total of 11 different types of fault datasets for each of the cases of normal behavior using multiple parameters and randomized runs to help an AI model recreate normal behavior.

\subsection{Results obtained so far: Data Representation}
The testbed simulates engine behavior over a 30-minute cycle. The data recorded during the simulation consist of the engine speed and torque, recorded once per second, and the control signals, the internal engine state, and the sensor output, recorded at intervals throughout the 30-minute cycle. To simulate engine faults, the simulation injects faulty data into the simulation pipeline (illustrated in Figure~\ref{fig:our_approach}) at one of the 3 locations depending on the type of fault. The linear interpolation technique was used to generate fixed-length sequences for each simulation. For each faulty class and non-faulty class concerning WLTP, data were recorded for each class with a thousand simulations which consisted of data with five files.\\
\textit{Omega} - reference engine speed.\\
\textit{Torque} - reference engine torque.\\
\textit{input\_signal} - 5 actuator measurements to the engine (control inputs for throttle position area and wastegate, engine speed, ambient temperature and pressure)\\
\textit{output\_signal} - 9 sensor measurements from the engine (temperatures for the compressor, intercooler and intake manifold, pressures for the compressor, intercooler, intake manifold and exhaust manifold, air filter mass flow and engine torque).\\
\textit{states\_signal} - 13 states of the engine system (temperatures and pressures for the air filter, compressor, intercooler, intake manifold, exhaust manifold, and turbine and turbine speed).

Given a sequence of control signals and sensor data from a potentially faulty engine, the aim is to predict the existence of a fault, if any, as early as possible in the sequence. To do so, I have initially implemented a Recurrent Neural Network and then a Transformer model.

\subsection{Challenges with the datasets and faults behavior}
The challenges with these time series datasets are the following: the number of rows and columns is not consistent in all five files of each simulation data. Also, each fault has a different behavior in terms of the starting point and ending point, and some of them are pulse faults, and some of them are abrupt faults in nature. In other words, some of them are periodic, some of them are periodic and consistent, etc.

\subsection{Recurrent Neural Network (RNN)}
My initial approach involves using a Recurrent Neural Network (RNN) to detect faults. In this experiment, I use a stacked RNN with 10 layers and 512 hidden nodes per layer followed by a fully-connected layer to produce a classifier with 12 output heads (one representing fault-free data and one for each of the 11 fault types). The RNN outputs a sequence of fault-type predictions for each time step. I train the RNN to reproduce the fault signal generated by the estimator/observer from the testbed; concretely, to train the network, I minimized the cross-entropy loss between the RNN output sequence and the fault signal sequence. This model was tested on one V100 Nvidia GPU accelerator, CPUs with 40 cores, 384GB RAM, and achieved 62.85\% accuracy on a holdout test set on the UTSA Arc HPC environment.

\section{Transformer Model} 
I designed a Transformer-based NN model for a diesel engine time series data set to perform fault classification and prediction. 
\subsection{Model architecture}
In this experiment, the model used 27 input dimensions,  64 hidden dimensions with 2 layers, and 9 heads to produce a classifier with 12 output heads (one representing fault-free data and one for each of the 11 fault types). I trained the model and minimized the cross-entropy loss between the Transformer output sequence and the fault signal sequence. This Transformer model with 27 parameters is trained on a system containing 4 NVIDIA GPUs V100, CPUs with 40 cores, 384GB RAM and this model achieves 70.01\% accuracy on a held-out test set for 20 epochs on UTSA Arc HPC environment. This experiment was also tested on five GPU nodes, each containing two CPUs with 20 cores each for a total of 40 cores, 384GB RAM, and each including two V100 Nvidia GPU accelerators and achieved 70.09\%. I analyzed the accuracy and time taken for each epoch and realized that there are not much differences in terms of accuracy, but after the seventh epoch, training was faster when submitted in batch mode. I have also tested this experiment on a large memory node, equipped with two AMD EPYC CPUs and 2 TB of RAM – 1 node equipped with two AMD EPYC CPUs and having 1 TB of RAM and found the almost same accuracy with 20 epochs.

 Input files consist of 27 parameters like engine speed, torque, 5 actuator measurements to the engine (control inputs for throttle position area and wastegate, engine speed, ambient temperature, and pressure), 9 sensor measurements from the engine (temperatures for the compressor, intercooler, and intake manifold, pressures for the compressor, inter-cooler, intake manifold, and exhaust manifold), air filter mass flow, and engine torque) and 13 states of the engine system (temperatures and pressures for the air filter, compressor, intercooler, intake manifold, exhaust manifold, turbine, and turbine speed). The output of the model is 12 classes (11 different types of faulty and non-faulty data).

To use this model, I converted the time series data into a sequence of fixed-length vectors or "tokens", which further fed into the Transformer model. One approach for using Transformers with time series data is to use a sliding window to create overlapping segments of the time series data and treat each segment as a separate "token". This creates a sequence of fixed-length vectors that can be fed into the Transformer model. The key feature of the Transformer architecture is its attention mechanism, which allows the model to selectively focus on different parts of the input sequence and learn complex patterns.
\subsection{Performance of Transformer Model}
\begin{figure}[h]
  \centering
  \includegraphics[width=\linewidth]{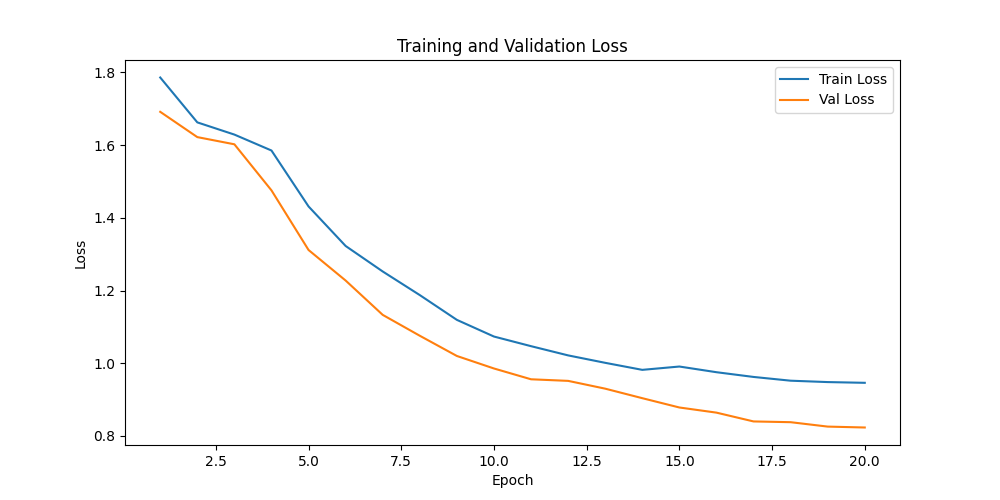}
  \caption{Training loss and validation loss}
\end{figure}

\begin{figure}[h]
  \centering
  \includegraphics[width=\linewidth]{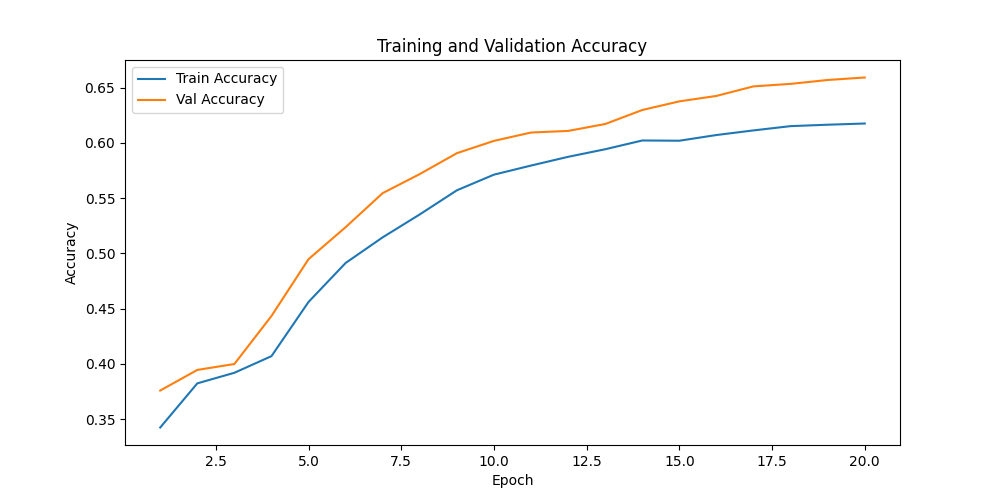}
  \caption{Training accuracy and validation accuracy}
\end{figure}

\subsection{Algorithms}
\begin{algorithm}
\caption{Data Preprocessing}
\label{alg:preprocess}
\begin{algorithmic}[1]

\Function{ReadFile}{$path, folder, fault$}
\State Read input, output, state data
\State Fix input missing values

\State Add rows in other data

\State Merge data, add fault label
\State \Return merged dataframe
\EndFunction

\Function{Preprocess}{$fault_paths$}
\State $combined \gets$ empty dataframe
\For{$fault\_path$ \textbf{in} $fault\_paths$}
\For{$folder$ \textbf{in} $fault\_path$}
\State $df \gets$ \Call{ReadFile}{$fault\_path, folder, fault$}

\State $combined \gets$ concatenate$(combined, df)$
\EndFor
\EndFor
\State Extract fault labels
\State Convert to tensors
\EndFunction

\end{algorithmic}
\end{algorithm}

\begin{algorithm}
    \caption{CustomDataset Class}
    \label{alg:custom_dataset}
    \begin{algorithmic}[1]
        \State \textbf{class} CustomDataset(\textit{Dataset}):
        \State \quad \textbf{function} Initialize(self, merged\_tensor, fault\_tensor):
        \State \quad\quad \textit{self.merged\_tensor} $\gets$ \textit{merged\_tensor}
        \State \quad\quad \textit{self.fault\_tensor} $\gets$ \textit{fault\_tensor}
        \State
        \State \quad \textbf{function} Size(self):
        \State \quad\quad \textbf{return} \textit{size of} \textit{self.merged\_tensor}
        \State
        \State \quad \textbf{function} GetSample(self, idx):
        \State \quad\quad \textit{merged\_row} $\gets$ \textit{self.merged\_tensor[idx]}
        \State \quad\quad \textit{fault\_label} $\gets$ \textit{self.fault\_tensor[idx]}
        \State \quad\quad \textbf{return} \textit{merged\_row}, \textit{fault\_label}
    \end{algorithmic}
\end{algorithm}

\begin{algorithm}
    \caption{EncoderDecoderTransformer Class}
    \label{alg:encoder_decoder_transformer}
    \begin{algorithmic}[1]
        \State \textbf{class} EncoderDecoderTransformer(\textit{nn.Module}):
        \State \quad \textbf{function} Initialize(self, input\_dim, output\_dim, hidden\_dim, num\_layers, num\_heads, dropout):
        \State \quad\quad \textbf{super}(\textit{EncoderDecoderTransformer, self}).\textbf{Initialize}()
        \State
        \State \quad\quad \textit{self.transformer} $\gets$ \textit{nn.Transformer}(
        \State \quad\quad\quad \textit{d\_model} $\gets$ \textit{input\_dim},
        \State \quad\quad\quad \textit{nhead} $\gets$ \textit{num\_heads},
        \State \quad\quad\quad \textit{num\_encoder\_layers} $\gets$ \textit{num\_layers},
        \State \quad\quad\quad \textit{num\_decoder\_layers} $\gets$ \textit{num\_layers},
        \State \quad\quad\quad \textit{dim\_feedforward} $\gets$ \textit{hidden\_dim},
        \State \quad\quad\quad \textit{dropout} $\gets$ \textit{dropout}
        \State \quad\quad )
        \State
        \State \quad\quad \textbf{function} Forward(self, merged\_tensor):
        \State \quad\quad\quad \textit{transformed} $\gets$ \textit{self.transformer}(merged\_tensor, merged\_tensor)
        \State \quad\quad\quad \textit{output} $\gets$ \textit{self.fc}(transformed)
        \State \quad\quad\quad \textbf{return} \textit{output}
    \end{algorithmic}
\end{algorithm}

\begin{algorithm}
\caption{Model Training and Evaluation}
\label{alg:train-eval}
\begin{algorithmic}[1]

\Function{Train}{$model,criterion,optimizer,train_loader$}
\State $model$.train()

\For{$data,target$ \textbf{in} $train_loader$}

\State $optimizer$.zero\_grad()
\State $data, target \gets data.to(device), target.to(device)$
\State $output \gets model(data)$
\State $loss \gets criterion(output, target)$
\State $loss$.backward()
\State $optimizer$.step()

\State Update $train\_loss, train\_correct$
\EndFor
\State \Return $train\_loss, train\_accuracy$
\EndFunction

\Function{Evaluate}{$model,criterion,val\_loader$}
\State $model$.eval()
\For{$data,target$ \textbf{in} $val\_loader$}
\State $data, target \gets data.to(device), target.to(device)$
\State $output \gets model(data)$
\State $loss \gets criterion(output, target)$
\State Update $val\_loss, val\_correct$
\EndFor
\State \Return $val\_loss, val\_accuracy$
\EndFunction

\end{algorithmic}
\end{algorithm}
\pagebreak
\section{Conclusion and Future Scope}
I have designed and implemented the RNN model with an accuracy of 62.85\% and the Transformer model on the UTSA Arc HPC environment which achieved 70.01\% accuracy on the simulated engine dataset. This work is still in progress and I will be working on improving the model accuracy. I will also test this model with the real-time dataset and find the model's accuracy. In the future, this can be experimented with with the Koopman NN and perform the performance analysis among these models. This could be also tested on the distributed learning environment HORVOD for the inferencing.

\section{Data Availability}
The datasets generated and/or analyzed during this study are not publicly available because they are intended to be used for a future benchmark submission.

\section{Contributions}
This research work was carried out under the mentorship of Dr. Sumit Kumar Jha, PI of the grant ``Office of Naval Research/Grant Number - N000014-21-1-2332I'' and he provided the data set and funds; I designed the AI models and conducted the experiments on the UTSA Arc HPC environment and submitted a report on this. I also presented a poster on this work in PAW'23 (Postdoc Appreciation Week) at UTSA which is available in the UTSA digital library.

\section*{Acknowledgment}
I express my sincere appreciation to Dr. Sumit Jha for his support and to the ``Office of Naval Research (ONR)'' for providing the research grant that funded this project. This work received computational support from the UTSA Arc HPC environment of UTSA, operated by University Tech Solutions. Special thanks to the ViZLab and support team for their invaluable support. Lastly, I extend our thanks to the CS department and technical support team for providing us with infrastructure, support, and assistance. 


\bibliographystyle{acm}
\bibliography{sample-base}

\begin{thebibliography}{10}

\bibitem{electronics11131969}
{\sc Bai, H., Zhan, X., Yan, H., Wen, L., and Jia, X.}
\newblock Combination of optimized variational mode decomposition and deep transfer learning: A better fault diagnosis approach for diesel engines.
\newblock {\em Electronics 11}, 13 (2022).

\bibitem{balakrishnan2021stochastic}
{\sc Balakrishnan, K., and Upadhyay, D.}
\newblock Stochastic adversarial koopman model for dynamical systems, 2021.

\bibitem{SILVA201210977}
{\sc da~Silva, J.~C., Saxena, A., Balaban, E., and Goebel, K.}
\newblock A knowledge-based system approach for sensor fault modeling, detection and mitigation.
\newblock {\em Expert Systems with Applications 39}, 12 (2012), 10977--10989.

\bibitem{Jiang22}
{\sc Jiang, J., Li, H., Mao, Z., Liu, F., Zhang, J., Jiang, Z., and Li, H.}
\newblock A digital twin auxiliary approach based on adaptive sparse attention network for diesel engine fault diagnosis.
\newblock {\em Scientific reports 12}, 1 (2022), 675.

\bibitem{6710264}
{\sc Loureiro, R., Benmoussa, S., Touati, Y., Merzouki, R., and Ould~Bouamama, B.}
\newblock Integration of fault diagnosis and fault-tolerant control for health monitoring of a class of mimo intelligent autonomous vehicles.
\newblock {\em IEEE Transactions on Vehicular Technology 63}, 1 (2014), 30--39.

\bibitem{6144754}
{\sc Loureiro, R., Merzouki, R., and Bouamama, B.~O.}
\newblock Bond graph model based on structural diagnosability and recoverability analysis: Application to intelligent autonomous vehicles.
\newblock {\em IEEE Transactions on Vehicular Technology 61}, 3 (2012), 986--997.

\bibitem{9036118}
{\sc Ng, K.~Y., Frisk, E., Krysander, M., and Eriksson, L.}
\newblock A realistic simulation testbed of a turbocharged spark-ignited engine system: A platform for the evaluation of fault diagnosis algorithms and strategies.
\newblock {\em IEEE Control Systems Magazine 40}, 2 (2020), 56--83.

\bibitem{Ng2020}
{\sc Ng, M. K.~Y., Frisk, E., Krysander, M., and Eriksson, L.}
\newblock A realistic simulation testbed of a turbocharged spark-ignited engine system: A platform for the evaluation of fault diagnosis algorithms and strategies.
\newblock {\em IEEE Control Systems Magazine 40\/} (03 2020), 56--83.

\bibitem{raj2019towards}
{\sc Raj, S., Fernandes, S.~L., Michel, A., and Jha, S.~K.}
\newblock Towards ai-driven predictive modeling of gas turbines using big data.
\newblock In {\em AIAA Propulsion and Energy 2019 Forum\/} (2019), p.~4385.

\bibitem{scacchioli2006model}
{\sc Scacchioli, A., Rizzoni, G., and Pisu, P.}
\newblock Model-based fault detection and isolation in automotive electrical systems.
\newblock In {\em ASME International Mechanical Engineering Congress and Exposition\/} (2006), vol.~47683, pp.~315--324.

\bibitem{4711448}
{\sc Tang, L., Kacprzynski, G.~J., Goebel, K., Saxena, A., Saha, B., and Vachtsevanos, G.}
\newblock Prognostics-enhanced automated contingency management for advanced autonomous systems.
\newblock In {\em 2008 International Conference on Prognostics and Health Management\/} (2008), pp.~1--9.

\end{thebibliography}
\end{document}